\documentclass[10pt,twocolumn,letterpaper]{article}
\usepackage{iccv}
\usepackage{times}
\usepackage{epsfig}
\usepackage{graphicx}
\usepackage{subfigure}
\usepackage{amsmath}
\usepackage{amssymb}
\usepackage{color}
\usepackage{float}

\usepackage[pagebackref=true,breaklinks=true,letterpaper=true,colorlinks,bookmarks=false]{hyperref}

\iccvfinalcopy 

\def\iccvPaperID{4725} 

\definecolor{orange}{rgb}{1,0.5,0}

\ificcvfinal\fi
\begin{document}

\title{Real Time Visual Tracking using Spatial-Aware Temporal Aggregation Network}
\author{Tao Hu \quad Lichao Huang \quad Xianming Liu \quad Han Shen\\
Horizon Robotics Inc\\
{\tt\small \{tao,hu, lichao.huang, xianming.liu, han.shen\}@horizon.ai}
}

\maketitle

\newcommand{\lichao}{\textcolor{red}}

\begin{abstract}
More powerful feature representations derived from deep neural networks benefit visual tracking algorithms widely. However, the lack of exploitation on temporal information prevents tracking algorithms from adapting to appearances changing or resisting to drift.
This paper proposes a correlation filter based tracking method which aggregates historical features in a spatial-aligned and scale-aware paradigm.  
The features of historical frames are sampled and aggregated to search frame according to a pixel-level alignment module based on deformable convolutions. In addition, we also use a feature pyramid structure to handle motion estimation at different scales, and address the different demands on feature granularity between tracking losses and deformation offset learning. By this design, the tracker, named as Spatial-Aware Temporal Aggregation network (SATA), is able to assemble appearances and motion contexts of various scales in a time period, resulting in better performance compared to a single static image. Our tracker achieves leading performance in OTB2013, OTB2015, VOT2015, VOT2016 and LaSOT, and operates at a real-time speed of 26 FPS, which indicates our method is effective and practical. Our code will be made publicly available at \href{https://github.com/ecart18/SATA}{https://github.com/ecart18/SATA}.
\end{abstract}

\begin{figure}[ht]
\begin{center}
   \includegraphics[width=1.0\linewidth]{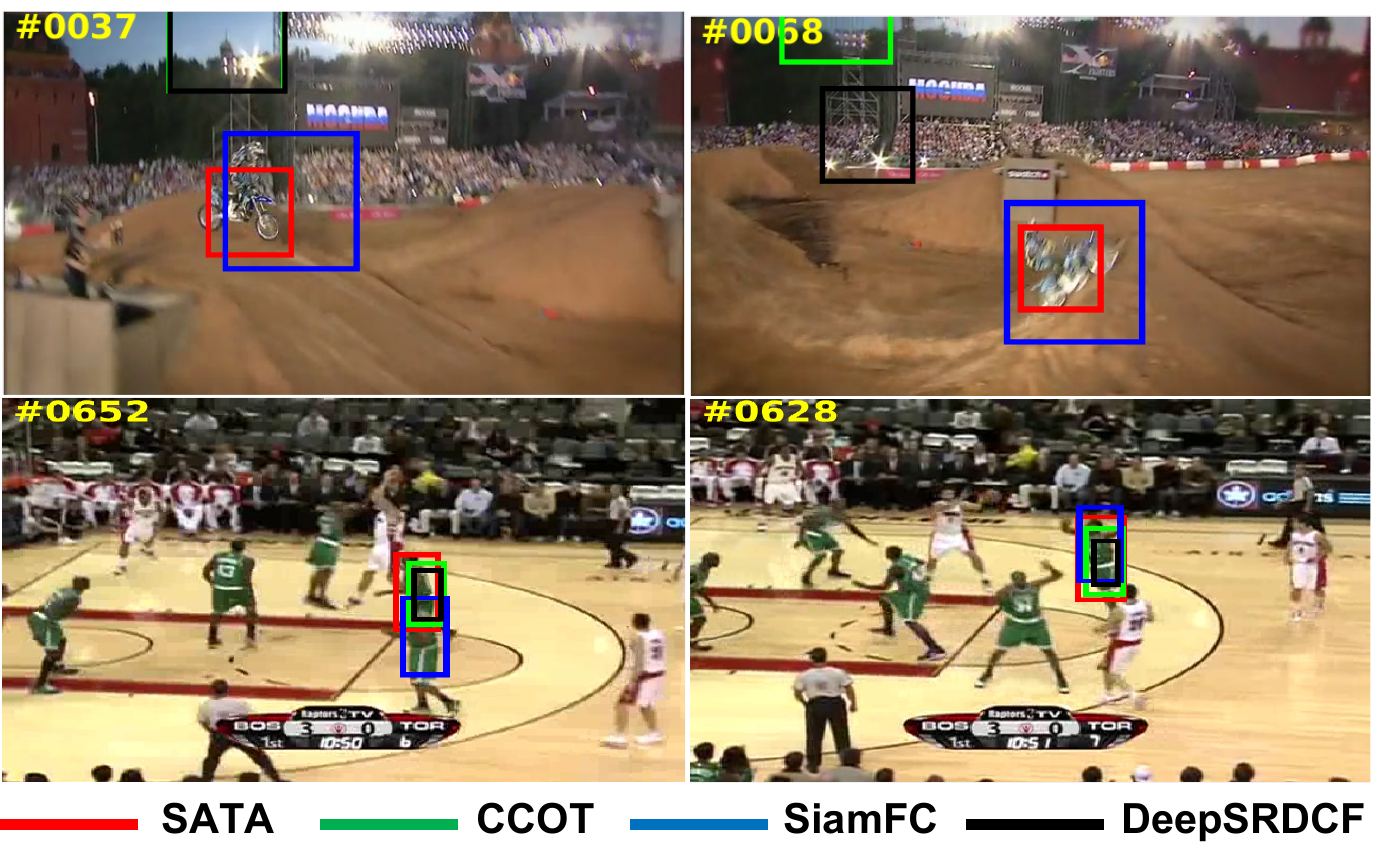}
\end{center}
\caption{Tracking results with the comparison to state-of-the-art deep DCF trackers including CCOT, SiamFC, and DeepSRDCF. Our SATA tracker enhances the localization robustness and accuracy of challenging scenarios.}
\label{fig.visual}
\vspace{-4mm}
\end{figure}

\section{Introduction}

As a fundamental problem in computer vision, visual object tracking, which aims to estimate locations of a visual target at each frame within a video sequence~\cite{song2018vital}, has been applied extensively on navigation, security and defense areas. Correlation filter based (CF) trackers have shown great potential on this task due to its impressive speed in the last decade. It attempts to get the best match position of objects from frame to frame and solves a ridge regression problem efficiently in Fourier frequency domain~\cite{henriques2015high}. Hand-crafted features are usually adopted in canonical CF trackers~\cite{henriques2015high, bolme2010visual, henriques2012exploiting, danelljan2014adaptive} until deep learning features become prevalent~\cite{danelljan2015convolutional, bhat2018unveiling}. To further improve the feature correlation, several fully convolutional network based correlation filter trackers have been proposed ~\cite{gundogdu2018good, wang2017dcfnet,valmadre2017end}, which demonstrated the superior performance and efficiency by combining a learnable feature extractor, such as CNN, with the correlation loss. Unfortunately, challenging scenarios such as partial occlusion, motion blur, and low-resolution remain unsolved with only considering the frame-level appearance and simple inter-frame position prior for most existing CF trackers \cite{lu2018visual}. 

\renewcommand{\thefootnote}{}\footnotetext{The work was done when Tao Hu was an intern in Horizon Robotics Inc.}

Visual tracking relies on motion and temporal context, which motivates us to incorporate more information from historical contents to improve the confidence of a tracker, analogy to using memories to enrich object representations in the current frame to improve tracking accuracy, especially when dramatic appearance changes such as rapid object motion and out-of-plane rotation. A more robust feature representation could be constructed by aggregating related historical frames. It has been exploited in video object detection recently~\cite{xiao2018video, zhu2017flow, bertasius2018object}, and brings noticeable performance improvements due to the better feature learned especially for challenging frames. To eliminate the appearance inconsistency introduced by object deformation, visual features need to be properly aligned before aggregation. 
The explored alignment methods include pixel-wise correlation~\cite{xiao2018video}, optical flow warping~\cite{zhu2017flow} or deformable convolution sampling~\cite{bertasius2018object}. 
Compared to applying an optical-flow based warping method~\cite{zhu2018end}, which requires intensive computation by extracting features in an extra branch and dense flow prediction~\cite{dosovitskiy2015flownet,ilg2017flownet}, we attempt to utilize deformable convolution networks~\cite{dai2017deformable} to estimate the motion transformation and conduct the spatial alignment in a more computational efficient manner.

In terms of scales of feature to align, shallow layers of CNN favor fine-grained spatial details and help locate precise object location~\cite{bhat2018unveiling, li2016deeptrack}, while deeper layers maintain semantic patterns that are robust to variation and motion blur. Recent deep tracker~\cite{ma2015hierarchical} employs multiple convolutional layers in a hierarchical weighted ensemble of independent CF models. However, such naive feature ensemble still lacks full use of multi-level semantics.

In this work, we present a novel spatial-aware temporal feature aggregation network for visual object tracking as shown in Figure \ref{fig.overall.arch}. The intuition is to find out a pixel-level alignment among historical frames and explore an elegant fusion of multi-resolution feature maps. The features of different layers from historical frames are aligned to their corresponding feature layers of the template frame through Align Module, and then aggregated by Aggregation Module. The features from different layers are combined gradually top-down to produce final predicted CF features. Our network architecture is able to balance the need of tracking for shallow layer features from CF loss and the need of translation invariance for deep semantic features aroused by deformable offset learning. As illustrated in Figure \ref{fig.visual}, our method enhances the localization robustness and accuracy in scenarios of intra-class variation and motion blur.


The key contributions of this paper are summarized as follows:
\begin{itemize}
    \item We apply a pixel-wise "align-and-aggregate" temporal fusion paradigm to the visual tracking task. Without using extra annotated data, the method can exploit information from previous frames to enhance both accuracy and robustness of the tracking algorithm.

    \item We propose a sequential multi-level feature aggregation mechanism which balances spatial details and semantic invariant properly.
    
    \item  This network is trained end-to-end and significantly boosts the performance on VOT2015, VOT2016, OTB-2013, OTB-2015 and LaSOT. We conduct extensive ablation studies and show significant enhancement achieved by each component. Our code will be made publicly available.
\end{itemize}

\section{Related Work}
In this section, we give a brief overview on the CF based visual tracking and spatial-temporal fusion approaches related to our method.

\subsection{Discriminative Correlation Filter Based Tracker}

Discriminative correlation filter (DCF) based tracker is one of the most import methods in visual tracking for its impressive performance and speed~\cite{danelljan2015convolutional,danelljan2015learning}. 
In these methods, the filters are trained by minimizing the least-squares loss for all circular shifts of a training template patch, and the new location for object in the search patch are obtained through finding the maximize value in the correlation response map~\cite{bolme2010visual}. 
The matching formulation is simple which makes the features become rather important, and many researchers start from this aspect to optimize the tracking performance. For example, CN~\cite{danelljan2014adaptive} combines the Color Name and grayscale features to describe the object.
KCF~\cite{henriques2015high} uses multi-channel HOG feature maps to extend single-channel gray features.
Recently, deep neural networks like CNN have attached more and more attention to researchers in the field of visual tracking~\cite{danelljan2015convolutional}. ImageNet~\cite{deng2009imagenet} pre-trained deep CNN models are utilized in DCF based methods to extract the features to correlate~\cite{ma2015hierarchical}. 
However, the achieved tracking performance may not be sufficient because the pre-trained features are extracted independently with the correlation tracking process. To address this issue, several methods~\cite{gundogdu2018good, wang2017dcfnet,valmadre2017end, bertinetto2016fully} interpret CF as a layer of the fully convolutional network based on Siamese architecture and train it with video detection datasets~\cite{russakovsky2015imagenet} from scratch. The integration of DCF with features from CNNs has achieved exciting performance boosting~\cite{wang2018multi}. However, these approaches have an inherent limitation When dealing with challenging scenario such as illumination changes, motion blur in practice. 

\subsection{Feature Aggregation in Video Analysis}
Feature aggregation is proposed to utilize motion cues and improve performance for video analysis task such as video object detection and video object segmentation(VOS)~\cite{ballas2015delving, tripathi2016context}. There are two mainstream categories methods of feature aggregation: recurrent neural networks (RNNs) based approaches and spatial-temporal convolution based approaches. Ballas et al.~\cite{ballas2015delving} proposes convolutional gated recurrent units (ConvGRU) which model the video sequence in spatial and temporal dimension simultaneously. This network shows the significant advantage of capturing long-distance dependencies and makes remarkable improvements in video object detection tasks~\cite{xiao2018video}. 

Another direction to fuse the motion dynamic across frames is the spatial-temporal convolution-based methods. Such methods have been widely used in VOS, which is defined as a one-shot video object learning problem similar to visual object tracking. CRN~\cite{hu2018motion} takes the inter-frame motion as  a guidance to generate an accurate segmentation based on cascaded refinement network. OSMN~\cite{yang2018efficient} uses a spatial modulator to integrate historical object location information in its segmentation framework. By introducing a distance transform layer, MoNet~\cite{xiao2018monet} separates motion-inconstant feature in historical frames to refine segmentation results. FGFA~\cite{zhu2017flow} aggregates temporal features by using weighted summation directly. However, as a typical video analysis task, visual object tracking has been less benefited from these advances in temporal feature aggregation. In fact, given more temporal context, it provides the potential to improve the robustness of the tracking algorithm. Besides, it is also consistent with the pattern of how human eyes track objects. These intuitions drive us to employ temporal feature aggregation in the visual tracking task.
 
\subsection{Motion Estimation by Deep Learning}
Motion estimation in videos requires correspondences in pixel-level of raw image or features-level to find the relationship between consecutive frames~\cite{tu2019survey}. Optical flow is widely used to estimate precise per-pixel localization between two input frames. Early attempts for estimating optical flow have focused on variational and combinatorial matching approaches~\cite{tu2019survey}. Recently, many works have shown that optical flow estimation can be solved as a supervised learning problem through CNNs~\cite{dosovitskiy2015flownet,ilg2017flownet,ranjan2017optical}. However, such flow estimation networks are always integrated as a bypass branch, which makes the training procedure more challenging when the task loss is not consistent with the flow loss. In addition, training a flow network usually requires large amounts of flow data, which is difficult and expensive to obtain. Instead, some recent works find the correspondences implicitly, without the need for extra annotations to supervise the model. STMN~\cite{xiao2018video} proposes a pixel-wise correlation module to learn the object motion between features of neighboring frames. STSN~\cite{bertasius2018object} uses several stacked deformable sampling modules to align the high-level features. Nevertheless, the tasks of detection and tracking are quite different, such that when high-level features work on detection tasks whereas not the case on tracking task because the latter often addresses the appearance details.

The work most related to ours in visual tracking is FlowTracker~\cite{zhu2018end}. It proposes a temporal feature aggregation method based on optical flow. Though good performance achieved, their network is quite inefficient because it trains and extracts the optical flow field on the raw image.
In addition, this method operates on fixed-length temporal windows, which has difficulty modeling variable motion duration and the long-term temporal information.

\begin{figure*}
\begin{center}
    \includegraphics[width=0.8\linewidth]{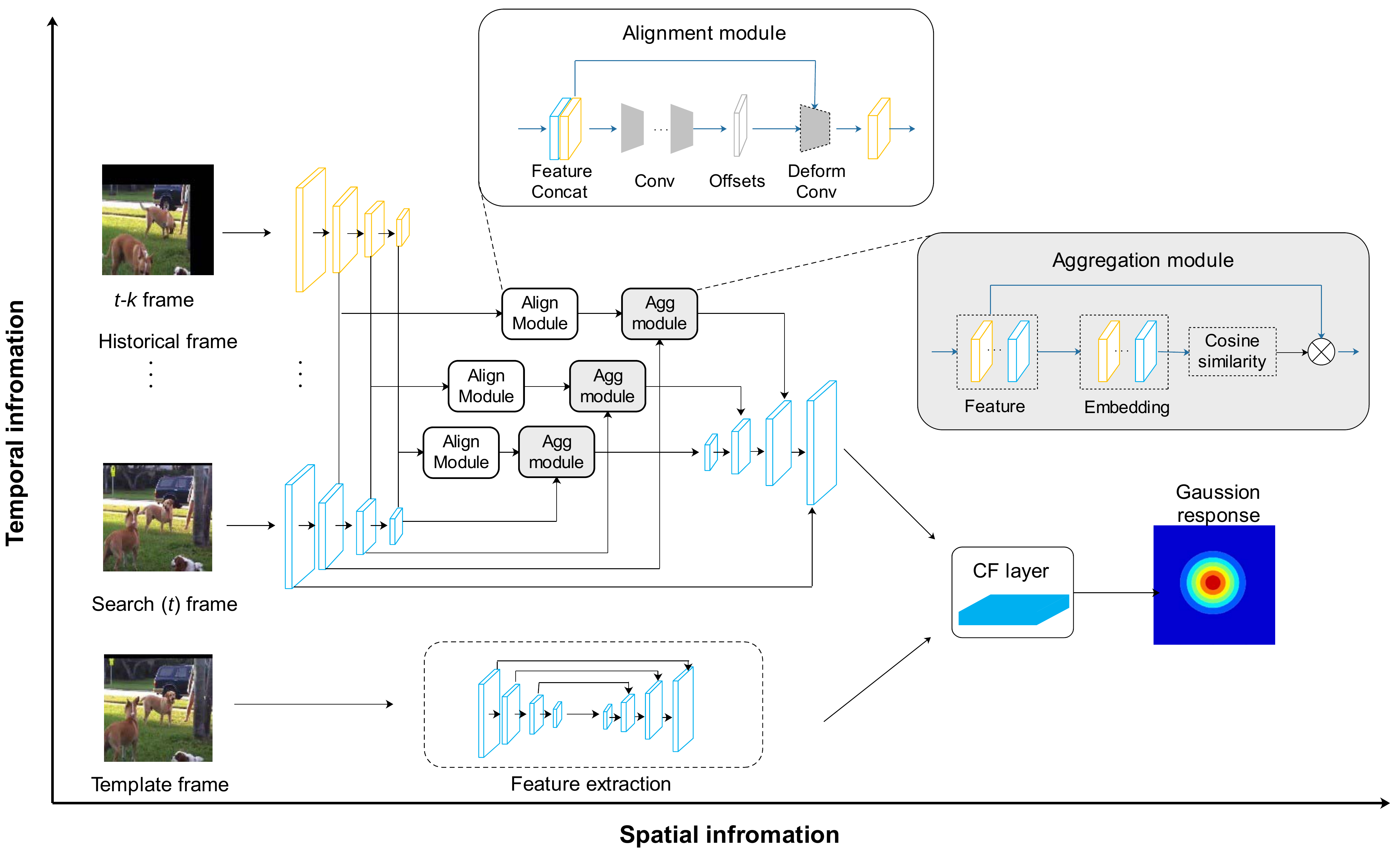}
\end{center}
   \caption{The overall of our SATA architecture. The network adopts Siamese architecture consisting of template branches and search branches. In search branches, the feature of historical frames are warped to search frame( $t$ frame). The white and grey boxes in middle part represent Alignment Module and Aggregation Module respectively. The feature map after aggregation of search frame and the feature map of template frame are fed to CF layer (right part).}
\label{fig.overall.arch}
\vspace{-4mm}
\end{figure*}

\section{Spatial-Aware Temporal Aggregation Network}

In this section, we first describe the overall architecture of this spatial-aware temporal aggregation network. Then we introduce each individual module in detail including the Alignment Module, the Aggregation Module and correlation filter layer.

\subsection{Multi-scale Feature Aggregation Mechanism}
\label{sec:msfam}
In a deep neural network like CNN, features of different layers show different patterns~\cite{yosinski2015understanding}. Generally, low level features contain sufficient appearance details, which has present better performance in localization, especially for visual tracking~\cite{gundogdu2018good,danelljan2015learning}. 
Respectively, the high layer features present semantic patterns to deal with the intra-class variation, which play an important role when the context varies with time.
To take advantage of the feature in both shallow and deep layers, we propose a multi-scale feature aggregation mechanism to learn the alignment across frames adaptively inspired by FPN ~\cite{lin2017feature}. 

As shown in Figure \ref{fig.overall.arch}, our framework adopts the Siamese network architecture, which consists of search and template branches. Given the input video frames $\left\{ I_t \right\}, t= 1,….,N $, we define the search frame index and its feature as $t$ and $F_t$ respectively, and the historical frame feature as $F_{t-\tau}$, where $\tau = 1,2...T$ is a integer large than zero in online tracking. Template frame, which is different from the history frames and the search frame, is selected randomly near the search frame.

In the search branch, we construct a feature pyramid for the search frame $I_t$ and historical frame $I_{t-\tau}$ through the backbone feature extraction network. 
Without loss of generality, we set the number of pyramid layers to $ l=3$, where $l$ represents the deepest layer of the feature pyramid for aggregation.
The feature pyramid consists of three proportionally sized feature maps, named $\left\{ P^{l}_{t}, P^{l-1}_{t}, P^{l-2}_{t} \right\}$ and $\left\{ P^{l}_{t-\tau}, P^{l-1}_{t-\tau}, P^{l-2}_{t-\tau} \right\}$ from deep layer to shallow layer for search and historical frame respectively.
Then, the Alignment Module (see details in section \ref{sec:fam}) are used to learn the relative offsets $\left\{ O^{l}, O^{l-1}, O^{l-2} \right\}$ for every feature map layer in $\left\{ P^{l}, P^{l-1}, P^{l-2} \right\}$ between $I_t$ and $I_{t-\tau}$. After that, $\left\{ O^{l}, O^{l-1}, O^{l-2} \right\}$ are used to generate the aligned feature $\left\{ P^{l}_{t-\tau \rightarrow t},  P^{l-1}_{t-\tau \rightarrow t}, P^{l-2}_{t-\tau \rightarrow t} \right\}$  with deformable ConvNets. Note that the Alignment Module is tailor-made for multi-scale pyramid feature to learn the motion dynamics and appearance change hierarchically. Finally, the $\left\{ P^{l}_{t-\tau \rightarrow t}, P^{l-1}_{t-\tau \rightarrow t}, P^{l-2}_{t-\tau \rightarrow t} \right\}$ are aggregated to$\left\{ P^{l}_{t}, P^{l-1}_{t}, P^{l-2}_{t} \right\}$ by temporal feature aggregation (see details in section \ref{sec:tfa}) and then merged through top-down pathway as FPN. 

In template branch, the feature maps of the template frame is extracted and the outputs of both branches are fed into the subsequent CF layer(see details in section \ref{sec:cfl}).

\subsection{Alignment Module}
\label{sec:fam}
A feasible method to improve frame feature quality is the temporal feature aggregation. However, directly aggregating feature always decline the model performance because of the features for the same object across frames are usually misaligned. To resolve feature misalignment issue, we design the Alignment Module which uses deformable convolution~\cite{dai2017deformable} to estimate the object motions and to achieve feature alignment across frames. To our best knowledge, this is the first attempt to apply deformable ConvNets for object tracking.

\noindent
\textbf{Feature Alignment procedure} 
For historical and search frames' feature $F_{t-\tau}$, $F_t$, we warp the $F_{t-\tau}$ to $F_t$ according to the Alignment Module, which consists of three major steps as illustrated in Figure \ref{fig.overall.arch} .

$Step1:$ The feature tensor $F_t$ and $F_{t-\tau}$ are of the same dimension that  $F_t, F_{t-\tau} \in R^{c \times w \times h }$, where $c, w, h$ denote the channel number, feature map width and height, respectively. We concatenate them along channels as $F_{t,t-\tau} \in R^{2c \times w \times h }$  to combine object appearance and motion information between the search and historical frames. 

$Step2:$ Next, to obtain the offsets of the historical frame w.r.t. the search frame for every pixel location in the feature map $F_{t-\tau}$, we use the concatenated feature tensor $F_{t,t-\tau}$ to predict an offsets $O_{t,t-\tau}$ with convolution layers. 

$Step3:$ Finally, the predicted offsets $O_{t,t-\tau}$ are used to sample the points in the feature map of historical frames $F_{t-\tau}$ via  deformable ConvNets. The output feature $F_{t-\tau \rightarrow t}$ are then used to aggregate with the feature of the search frame.

\noindent
\textbf{Tailor-made convolution network for offsets prediction} As described above, we use the concatenated feature tensor $F_{t,t-\tau}$ to predict the relative offsets by convolution network. However, the expected relative offsets are distinguishing for different spatial resolution features in feature pyramid. On the one hand, the shallow feature with large feature map size takes on large relative offsets. Hence, a large kernel size should be considered, but it will greatly increase the parameter number, bringing issues of over-fitting. To address this problem, we stack more convolution layers to extend the receptive field. On the other hand, the shallow features with edge and contour awareness are extremely fine-grained, which makes it difficult to predict the relative offsets. Therefore, we should design a larger capacity convolution network. On the contrary, a shallow and small convolution network are applicable robust for modeling the small offsets between high level feature. In practice, our Alignment Module has 2,3,3 convolution layers for features $P^{l}, P^{l-1}, P^{l-2}$ in feature pyramid respectively, and we use dilation convolution network to extend reception field with dilation 1.

In summary, the Alignment Module uses deformable ConvNets to learn the motion and appearance offsets between the historical frames and search frame.

\subsection{Temporal Feature Aggregation}
\label{sec:tfa}
The Alignment Module is applied for all selected historical frames in the specified range. 
After that, we obtain a series of complementary feature maps of historical frames which 
include diverse appearance information for the object being tracked. Inspired by the 
method in flow-guide feature aggregation network(FGFA)~\cite{zhu2017flow}, we use the adaptive weights 
to aggregate historical frames feature to search frames at each spatial location. 
The aggregation results could be formulated as:
\begin{equation}
F^{agg} = \sum_{\tau=1}^{T} w_{t-\tau \rightarrow t} F_{t-\tau \rightarrow t}
\end{equation}
Where $T$ is predefined range for historical frames and $w_{t-\tau \rightarrow t}$ is an adaptive weights mask for different spatial locations in the feature map $F_{t-\tau \rightarrow t}$. To compute the weight $w_{t-\tau \rightarrow t}$, we use a 3-layer fully convolutional bottleneck embedding network $E$ to compute the feature representation $e_{t-\tau \rightarrow t}$ for $F_{t-\tau \rightarrow t}$ firstly. Then cosine similarity are used to measure the similarity of each corresponding point location between $e_{t-\tau \rightarrow t}$ and search frame embedding feature $e_{t}$. We calculate the weights $w_{t-\tau \rightarrow t}$ by applying an exponential function on cosine similarity, and normalize all weights $w_{t-\tau \rightarrow t}$, $\tau \in \left[1,2, \cdots T \right]$ for all selected range of historical frames. The $w_{t-\tau \rightarrow t}$ could be formulated as:
\begin{equation}
w_{t-\tau \rightarrow t}(p) = \frac{\exp(\frac{e_{t-\tau \rightarrow t} \cdot e_{t}}{\left|e_{t-\tau \rightarrow t}\right| \cdot \left|e_{t}\right|})}{\sum_{\tau=1}^{T}{\exp(\frac{e_{t-\tau \rightarrow t} \cdot e_{t}}{\left|e_{t-\tau \rightarrow t}\right| \cdot \left|e_{t}\right|})}} 
\end{equation}
where $p$ is spatial location in feature map $F_{t-\tau \rightarrow t}$.

\subsection{Correlation Filter Layer}
\label{sec:cfl}
Discriminative correlation filter is one of the most popular solution for visual object tracking. Motivated by~\cite{gundogdu2018good, wang2017dcfnet, valmadre2017end}, we interpret correlation filter as a learnable layer embedded in the whole siamese tracking network.

In the standard DCF tracking framework for multi-dimensional feature, the goal is to learn series correlation filters $w^d, d \in \left\{1,2...D\right\}$ from training sample feature $F_x$ and the ideal response $y$. Here, feature tensor $F_x$, extracted from CNN, has $D$ channels and spatial size $m \times n$, and the ideal response tensor $y$ which is a Gaussian function peaked at the center has the same spatial size $m \times n$. The correlation filter $w^d$ could be obtained by minimizing the output ridge loss:
\begin{equation}
L = \left \| \sum_{d=1}^{D} w^d \star F_x^d - y \right \| ^2 + \lambda \sum_{d=1}^{D}\left \| w^d \right \|^2 
\end{equation}
Where $w^d$ and $F_x^d$ refers to the $d$th channel of filter $w$ and feature map of $x$ respectively, $\star$ denotes circular correlation operation. The $\lambda$ in the second term of equation is regularization coefficient. Theoretically, the solution can be formed as:
\begin{equation}
\hat{w_d} = \frac{\hat{F_x^d} \odot{\hat{y^\ast}}} {\sum_{d=1}^{D}\hat{F_x^d}  \odot{\left(\hat{F_x^d}\right)^\ast} + \lambda}
\end{equation}
where the $\land$ symbol and the $\ast$ symbol represents discrete Fourier transform and complex conjugate of a complex respectively, and the symbol $\odot$ denotes Hadamard products.

In the tracking process, we crop a search patch $z$ and obtain the corresponding feature tensor $F_z$, then the correlation response map $g$ for the search patch is calculated as:
\begin{equation}
g = \mathcal{F}^{-1} \left( \sum_{d=1}^{D} \hat{w}^{d\ast} \odot{\hat{F_z}^{d}} \right)
\end{equation}
Finally, the translation offsets could be obtained by searching the location of the maximum value of response map  $g$. In the Siamese framework, we train network parameters $w$ with template patch $x$, and calculate 
the loss with search patch $z$. The loss function for training is defined as:
\begin{equation}
l = \left \| g - y \right \| ^2 + \lambda \left \| \theta \right \|^2
\end{equation}
$ \theta $ represents parameters in CNN. Note that we embed correlation filters 
as a learnable layer for end-to-end training. The details of the derivation for back 
propagation could be found in ~\cite{gundogdu2018good, wang2017dcfnet}.

\section{Training \& Inference}
In this section, we detail the pipeline of the proposed SATA tracker for training and inference. 
\subsection{Training}
During the training phase, the input data fed to this siamese network architecture are clips of the video sequences, which consist of historical frames, search frame and template frame. More specifically, we randomly sample historical frames before the search frame, and randomly select another frame near the search frame as template frame. 
The backbone feature extraction network is applied on individual frame to produce the multi-scale feature maps. The Alignment Module in Section~\ref{sec:fam} is used to estimate the sampling offsets between the historical frames and the search frame, and the feature maps from historical frames are sampled to the search frame according to the predicted sampling offsets in multi-scale layer. The sampled feature maps, as well as its own feature maps on the search frame, are aggregated through Aggregation Module in Section~\ref{sec:tfa}. Finally, the resulting aggregated feature maps of the search frame and the feature of template frame are then fed to the correlation filter layer to produce the tracking result. 
To explore more feature data for training Alignment Module, data augmentations are adopted including affine transformation to history frames. All the modules in this framework are trained end-to-end.

\subsection{Inference}

After off-line training, the learned network is used to perform online tracking. We fix backbone feature extraction network and other CNN layer but update parameters in correlation filter layer during the online tracking. The first frame of a video is set as template frame in tracking. Given a new frame, we crop and resize a search patch centered at the estimated position in the last frame. The historical frames are selected from previous frames of the search image with fixed window size $T$. Then the feature pyramid of historical frames are warped and aggregated to search patch with Alignment Module and Aggregation Module respectively. The estimation location of the search patch is obtained by finding the maximum value in the response score map of CF layer. To make tracker adaptive to appearance variations, we update our CF layer network frame-by-frame as ~\cite{danelljan2014accurate}. To handle the scale changes, we carry out scale estimation followed the approach in ~\cite{danelljan2014accurate} and use scale pyramid with the scale factors $ \left\{ \alpha^s | s= \left \lfloor - \frac{S-1}{2} \right \rfloor, ... , \left \lfloor \frac{S-1}{2} \right \rfloor \right\} $.

\section{Experiments}

In this section, we first introduce the implementation details of experiments. Then, we perform in-depth analysis of SATA on various benchmark datasets including OTB-2013~\cite{wu2013online}, OTB-2015~\cite{wu2015object}, VOT-2015 and VOT-2016 ~\cite{kristan2015visual}. All the tracking results use the reported results to ensure a fair comparison. Finally, we present extensive ablation studies on the effect of spatial and temporal aggregation.

\subsection{Implementation details}
\noindent
\textbf{Training}
We implement the pre-trained VGG-16 model~\cite{simonyan2014very} as the backbone feature extraction network. Specifically, the output feature map of the Conv2, Conv4, Conv7, Conv10 layer are used to constructed feature pyramid. We apply $ 3 \times 3 \times 64 $ convolution layer as lateral to reduce and align the the channels of feature pyramid to 64. The feature map size for $ P^{1}, P^{2}, P^{3} $ are 62,31,15 respectively, The final output feature map size and response map size are both  $ 125 \times 125$.  Align Module implement as described in section \ref{sec:fam}. Embedding sub-network in spatial attention consists of three convolution layers ($ 1 \times 1 \times 32, 3 \times 3 \times 32, 1 \times 1 \times 64$ ). The lateral layer, Align Module and embedding sub-network are initialized randomly. The parameters in backbone feature extractor are fixed after pre-trained using ImageNet~\cite{deng2009imagenet}. We apply stochastic gradient descent (SGD) with momentum of 0.9 to end-to-end train the network and set the weight decay rate to 0.0005. The model is trained for 50 epochs with a learning rate of $10^{-5}$ and mini-batch size of 32. The regularization coefficient $\lambda$ for CF layer is set to 1e-4 and the Gaussian spatial bandwidth is 0.1.

\noindent
\textbf{Data dimensions}
Our training data comes from ILSVRC-2015 ~\cite{russakovsky2015imagenet}, which consist of training and validation set. For each training samples, we randomly pick a pair of image as template and search image within the nearest 10 frames. Historical frames are selected among the nearest 20 frames of the search frame randomly. The historical frame number of aggregation is set to 3 in training. In each frame, patch is cropped around ground truth with a 2.0 padding and resized into $ 125 \times 125 $. 

\noindent
\textbf{Inference} In online tracking, scale step a and number S is set to 1.03 and 3, scale penalty and model updating rate is set to 0.993 and 0.01 for scale estimation. Hyper-parameters in CF layer are seted as training. The learning rate for updating CF layer is 0.01. The proposed SATA-NET is implemented using Pytorch 4.0 ~\cite{paszke2017pytorch} on a PC with an Intel i7 6700 CPU, 48 GB RAM, Nvidia GTX TITAN X GPU. Average tracking speed of the SATA is 26 FPS and the code will be made publicly available.

\subsection{Results on OTB}
OTB is a standard and popular tracking benchmark with contains 100 fully annotated videos.  OTB-2013 contains 50 sequences, 
and OTB-2015 is the extension of OTB2013 which contains other 50 more difficult tracking video sequences. The evaluation of OTB is based on two metrics: center location error and bounding box overlap ratio. We follow the standard evaluation approaches of OTB and report the results based on success plots and precision plots of one-pass evaluation (OPE).
We present the results on both OTB-2013 and OTB-2015.

\begin{figure}[hbp]
    \begin{minipage}[t]{0.49\linewidth} 
    \centering 
    \includegraphics[width=1.\textwidth]{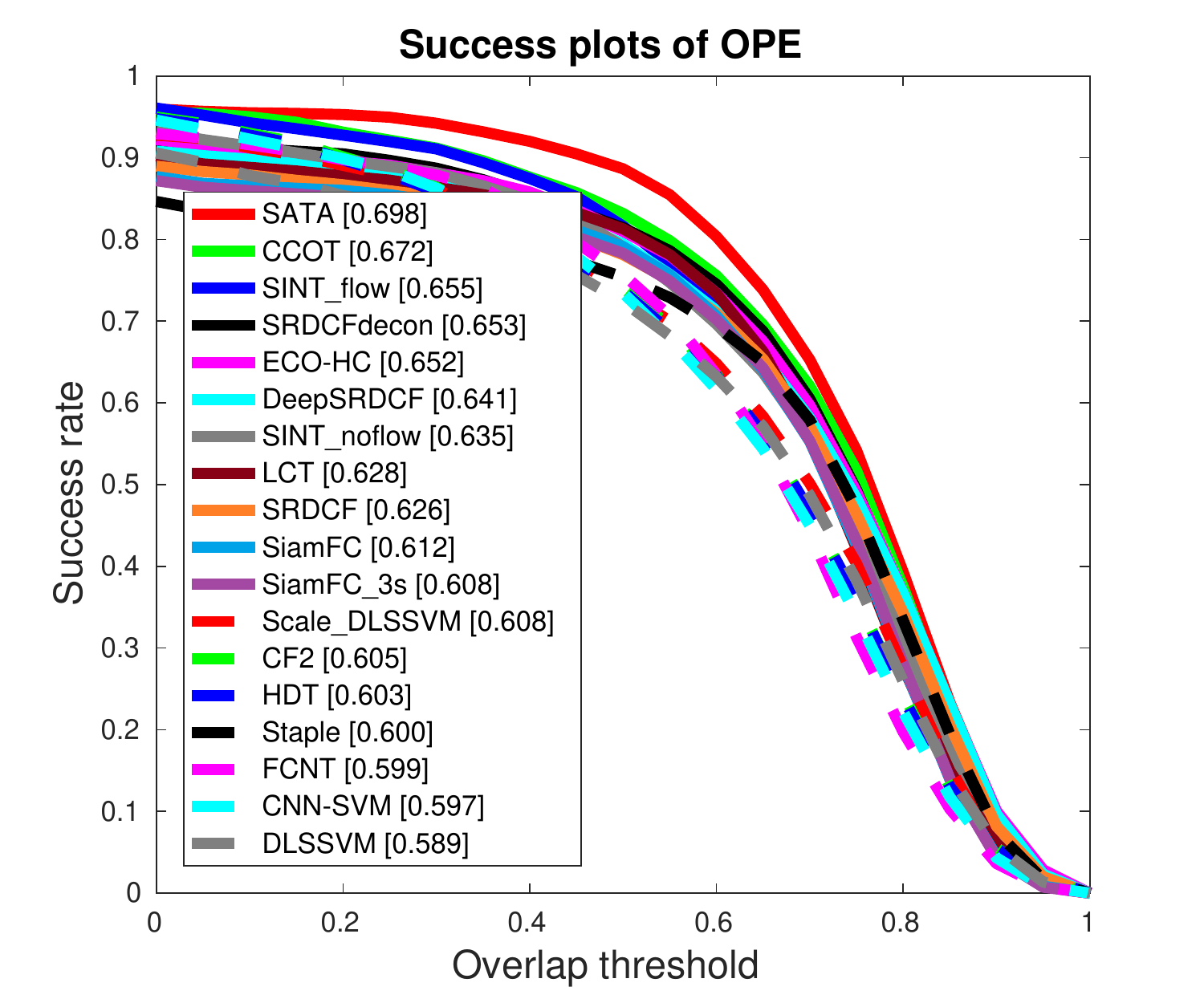} 
    \end{minipage}  
\hfill 
    \begin{minipage}[t]{0.49\linewidth} 
    \centering 
    \includegraphics[width=1.\textwidth]{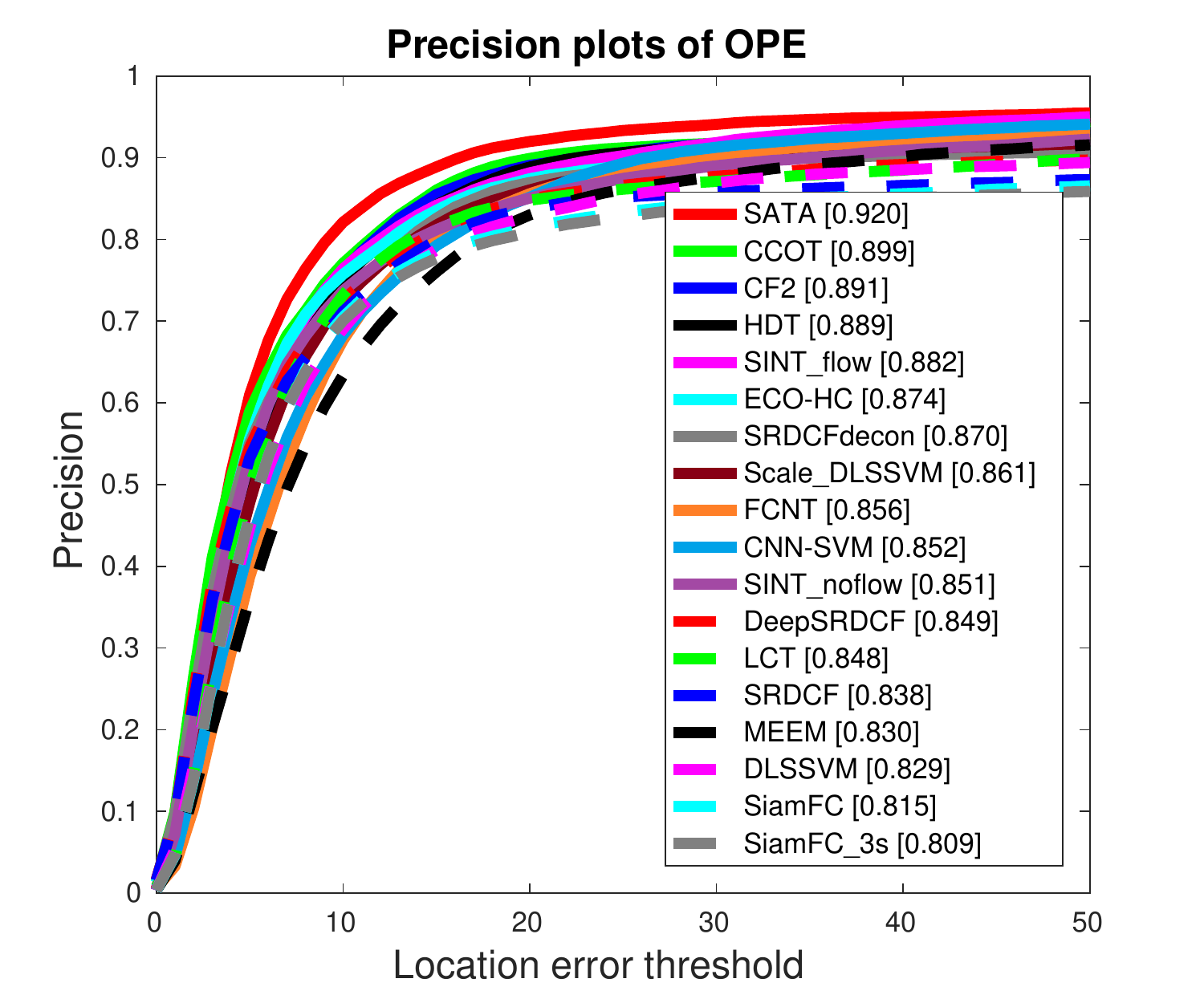}
    \end{minipage} 
\caption{Success and precision plots on OTB2013. The numbers in the legend indicate
the representative the area-under-curve scores for success plots and precisions at 20 pixels for precision plots.}
\label{otb13}
\vspace{-2mm}
\end{figure}

\begin{figure*}[htbp] 
\centering 
\subfigure[Low resolution]{ 
\begin{minipage}[t]{0.20\linewidth} 
\centering 
\includegraphics[width=1.0\textwidth]{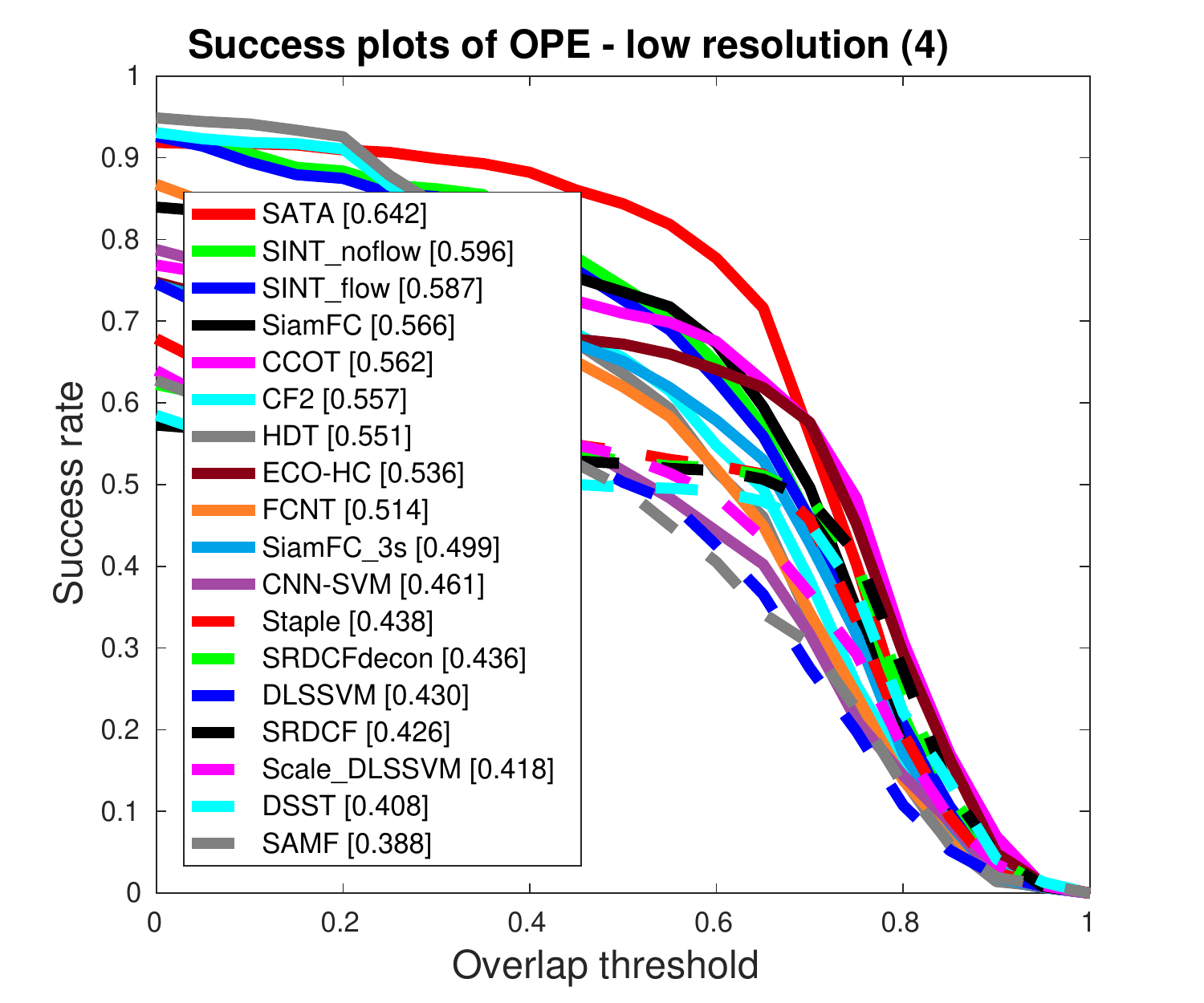}  
\end{minipage}  
}%
\subfigure[Illumination variations]{ 
\begin{minipage}[t]{0.20\linewidth} 
\centering 
\includegraphics[width=1.0\textwidth]{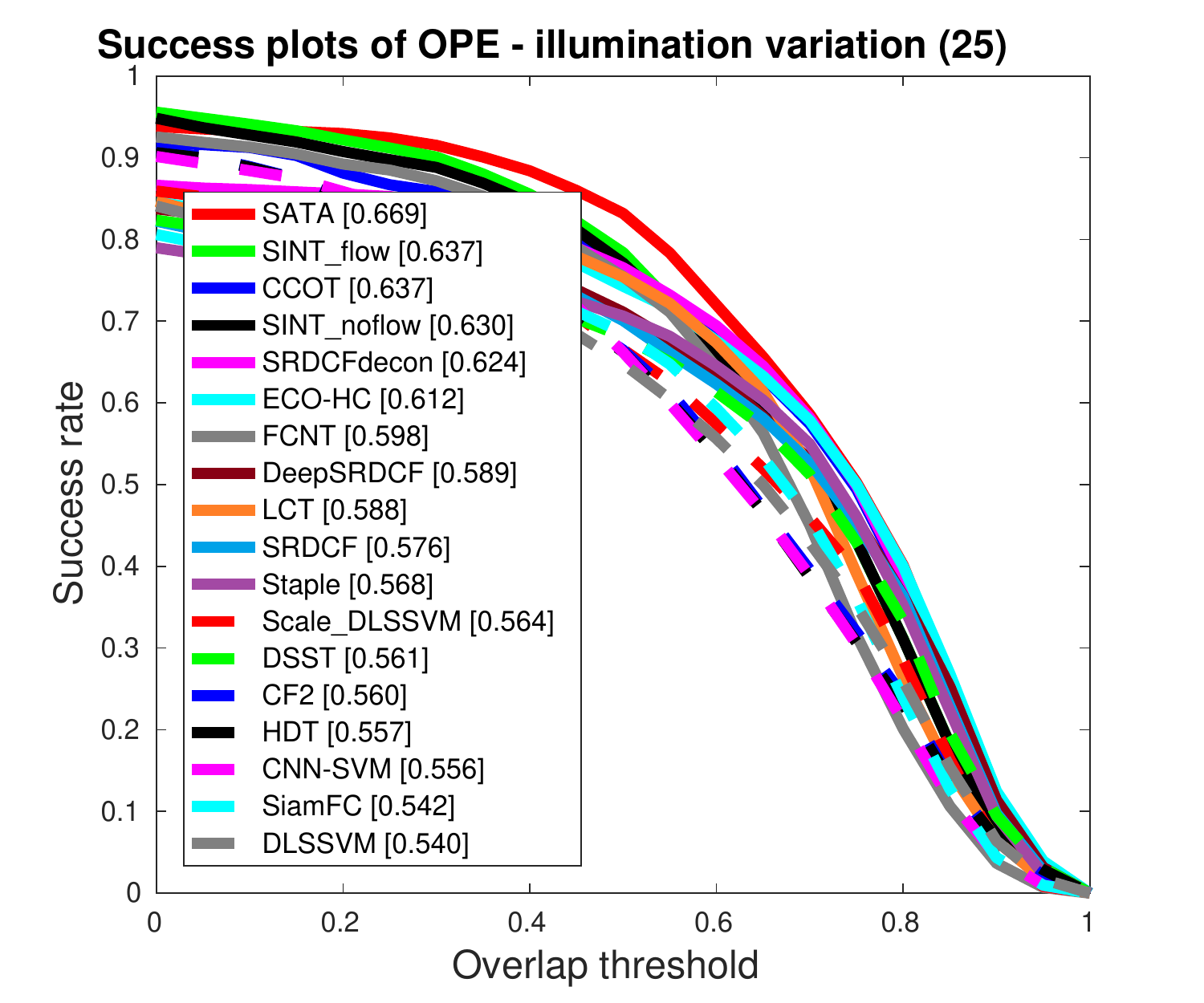} 
\end{minipage} 
}%
\subfigure[Scale variations]{ 
\begin{minipage}[t]{0.20\linewidth} 
\centering 
\includegraphics[width=1.0\textwidth]{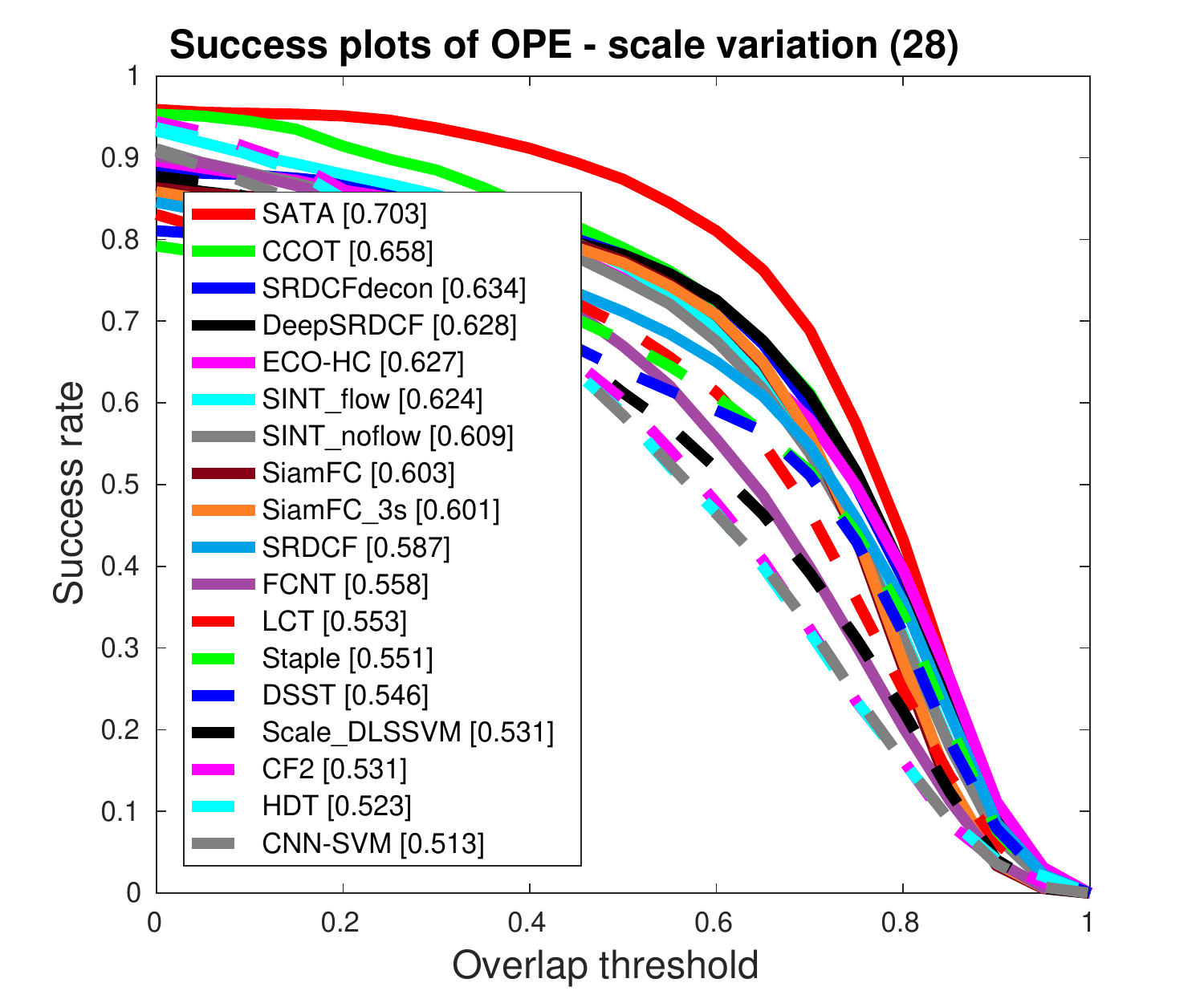} %
\end{minipage} }%
\subfigure[In plane rotation]{ 
\begin{minipage}[t]{0.20\linewidth} 
\centering 
\includegraphics[width=1.0\textwidth]{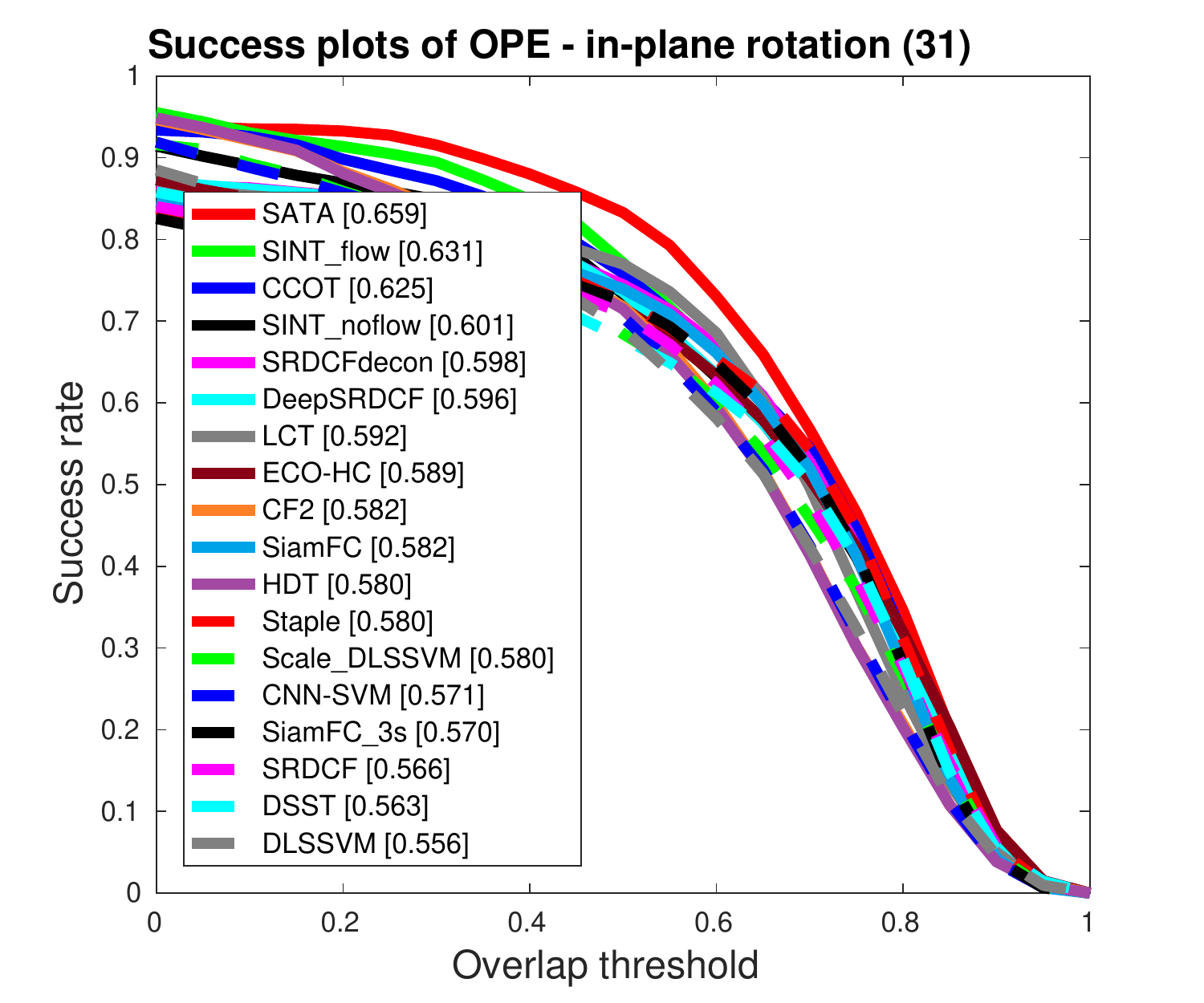} %
\end{minipage} }%
\centering 
\subfigure[Out of plane rotation]{ 
\begin{minipage}[t]{0.20\linewidth} 
\centering 
\includegraphics[width=1.0\textwidth]{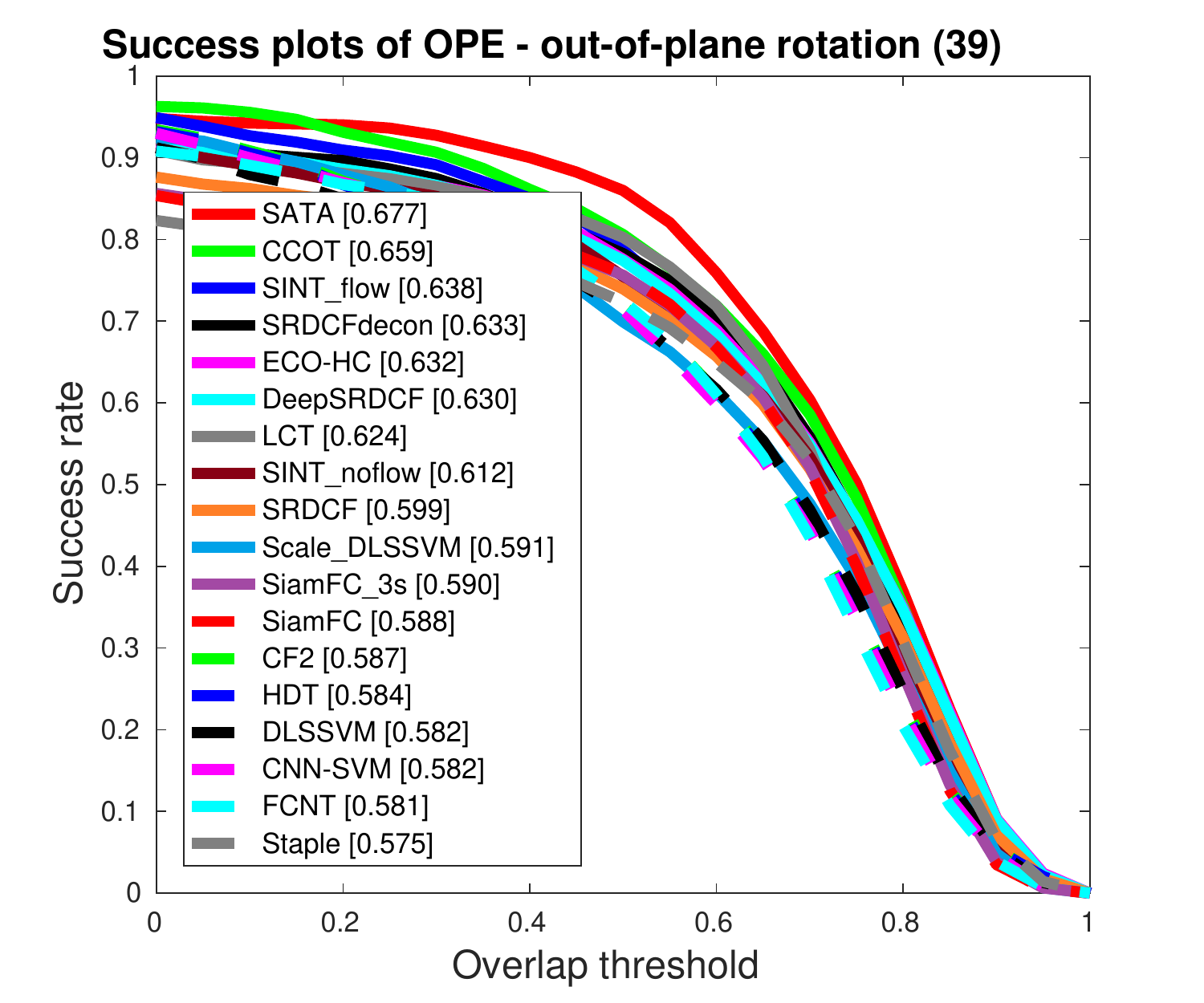} 
\end{minipage}  
}%
\centering 
\caption{Success plots with attributes on OTB2013}
\label{attri_otb13}
\vspace{-4mm}
\end{figure*}

\noindent
\textbf{OTB-2013} OPE is employed to compare our algorithm with the other 17 trackers from OTB-2013 benchmark, which including CCOT~\cite{danelljan2016beyond}, ECO-HC~\cite{danelljan2017eco}, DeepSRDCF~\cite{danelljan2015learning}, SiamFC~\cite{bertinetto2016fully}, et al. Figure \ref{otb13} shows the results from all compared trakcers, our SATA tracker performs favorably against state of art trackers in both overlap and precision success. In the success plot, our approach obtain an AUC score of 0.698, significantly outperforms the winner of VOT2016 (CCOT). In the precision plot, our approach obtains a highest score of 0.920.
For detailed performance analysis, we also report the performance under five video challenge attributes using one-pass evaluations in OTB2013
Figure \ref{attri_otb13} demonstrate that our SATA tracker effectively handles large appearance variations well caused by low resolution, illumination variations, in-plane and out-of-plane rotations while other trackers obtain lower scores.

\begin{figure}[htbp]
    \begin{minipage}[t]{0.49\linewidth} 
    \centering 
    \includegraphics[width=1.\textwidth]{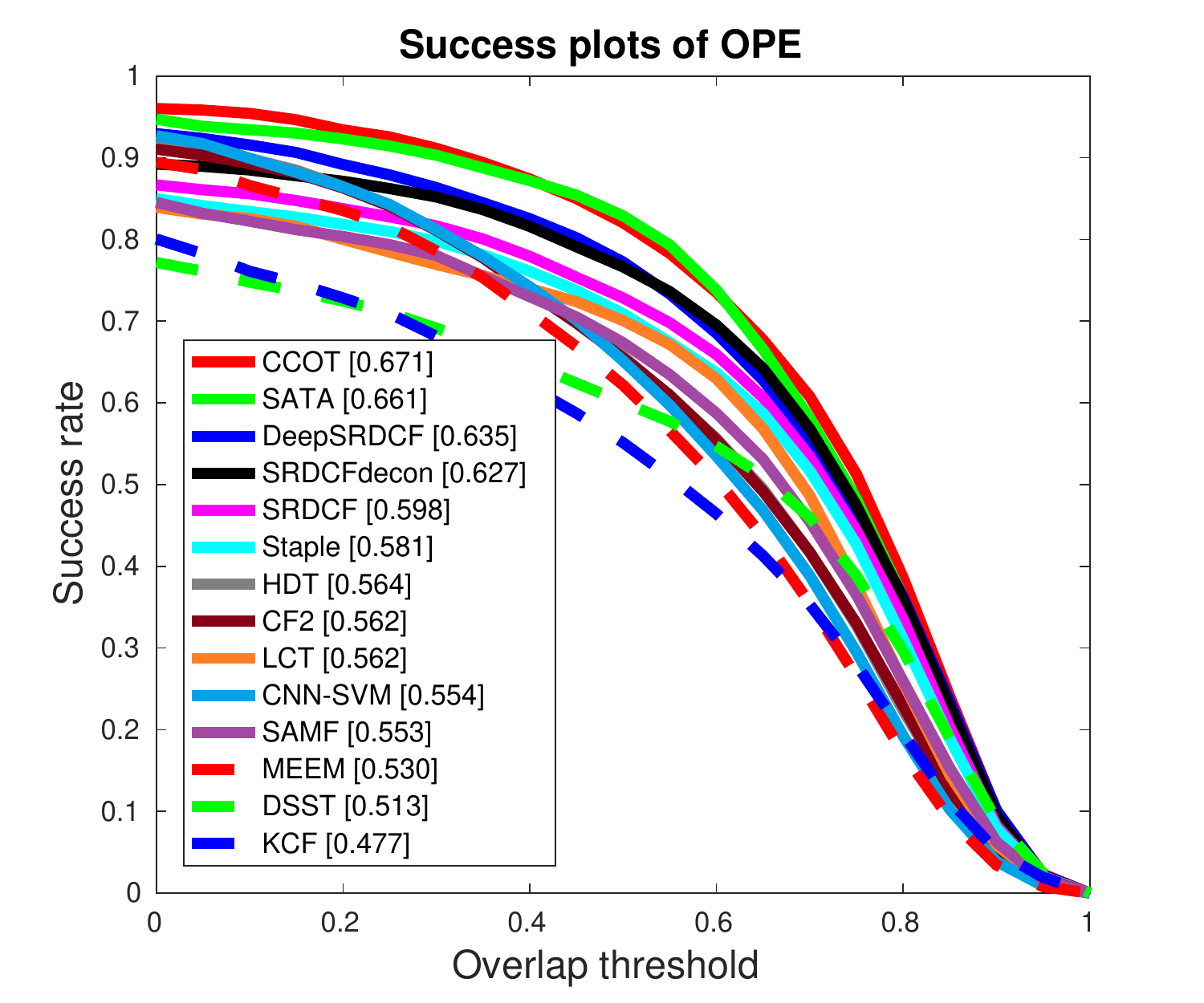} 
    \end{minipage}  
    \hfill 
    \begin{minipage}[t]{0.49\linewidth} 
    \centering 
    \includegraphics[width=1.\textwidth]{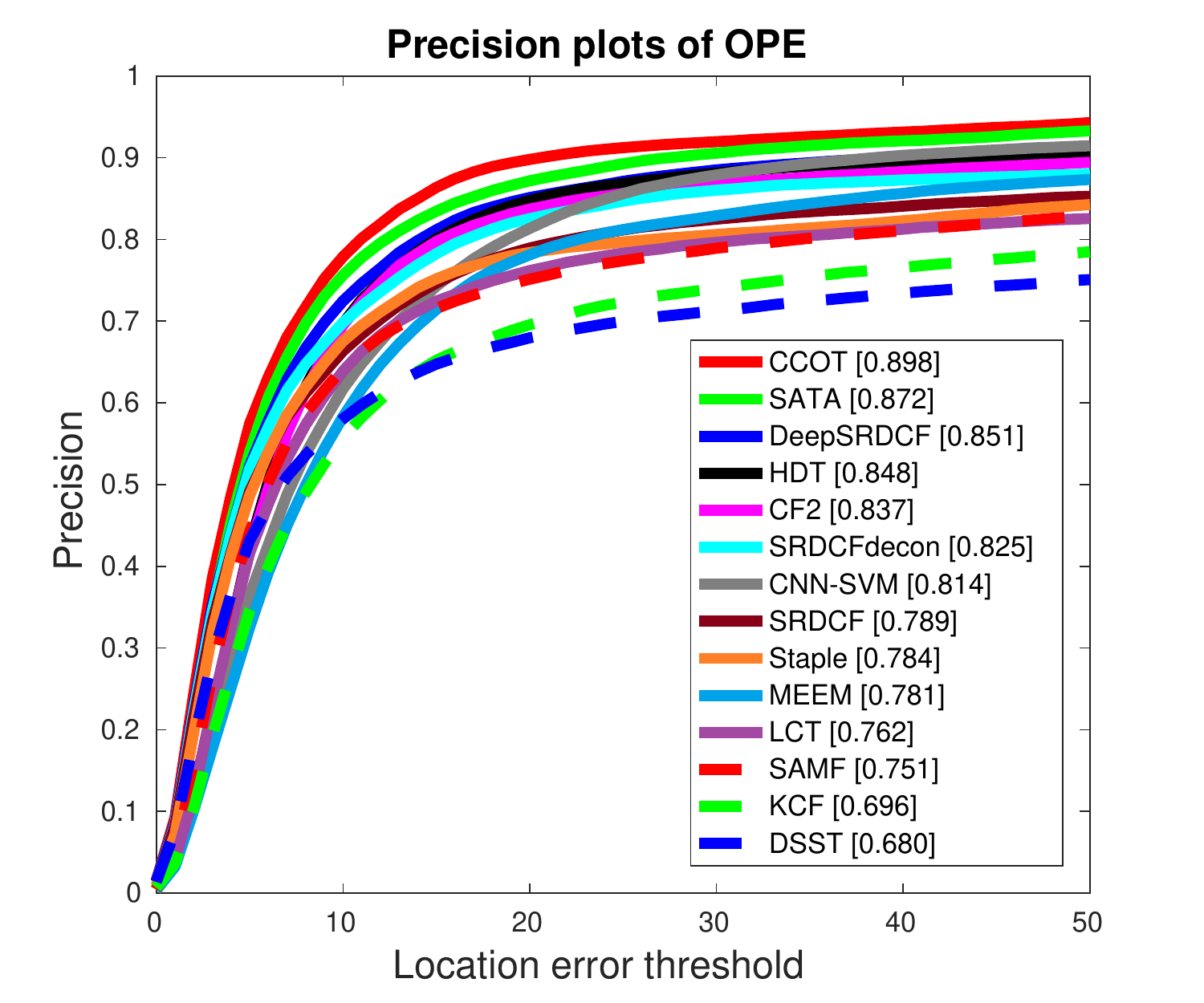}
    \end{minipage} 
\caption{Precision and success plots on OTB2015.}
\label{otb15}
\vspace{-4mm}
\end{figure}

\noindent
\textbf{OTB-2015} In this experiment, we compare our method against recent trackers on the OTB-2015 benchmark. Figure \ref{otb15} illustrates the success and precision plots of OPE respectively. The results shows that our SATA tracker overall performs well. Specifically, our method achieves a success score of 0.661, which is far superior to other methods combining correlation filter and deep learning such as DeepSRDCF(0.635) and CFNet(0.568) ~\cite{valmadre2017end}.

\subsection{Results on VOT}
For completeness, we also present the evaluation results on benchmark datasets VOT-2015 and VOT-2016, which both consists of 60 identical challenging videos but the ground truth has been re-annotated in VOT-2016.

\begin{figure}[h]
\begin{center}
   \includegraphics[width=0.9\linewidth, height=0.7\linewidth]{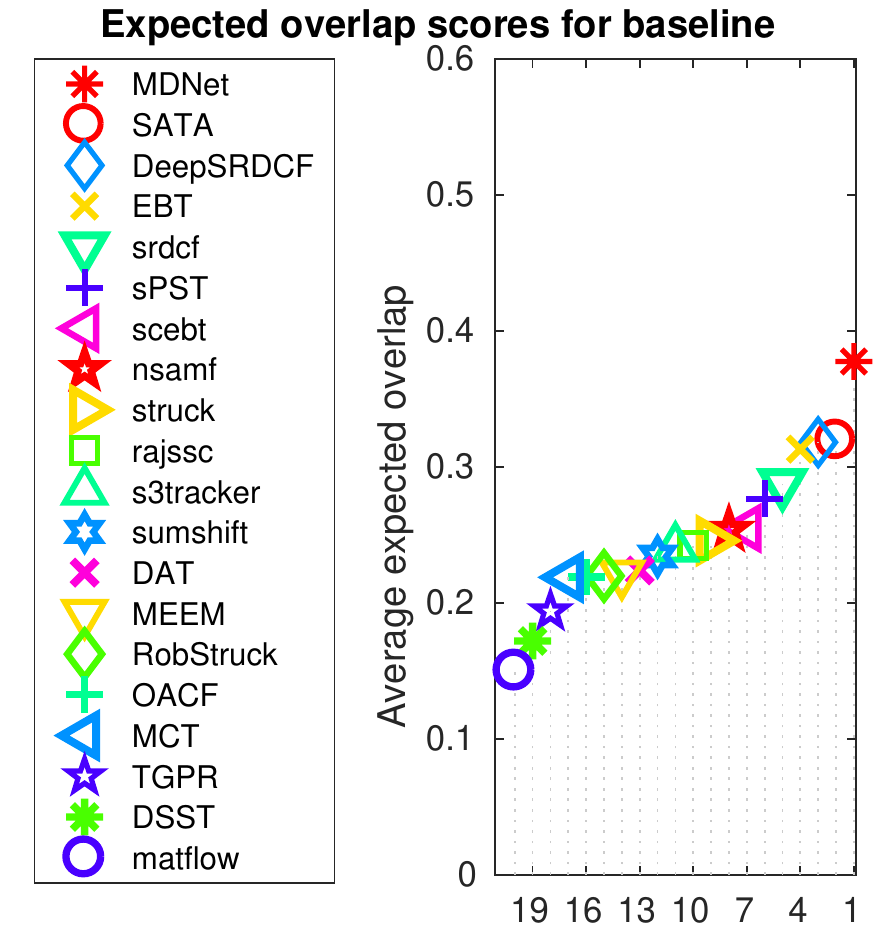}
\end{center}
   \caption{EAO ranking with the top 20 trackers in VOT-2015. The better trackers are located at the right.}
\label{fig.vot2015}
\end{figure}

In VOT2015, we present a state-of-the-art comparison to the top 20 participants in the challenge under the evaluation metric expected average overlap (EAO). Figure \ref{fig.vot2015} show that the performance of SATA ranks $2nd$ after MDNet~\cite{nam2016learning} only which is trained on OTB.

In VOT2016, we compare our SATA tracker with other state-of-the-art trackers including CCOT, MDNet and SiamFC. The performance of SATA is comparable to CCOT and better than SiamFC and MDNet. VOT-2016 report \cite{kristan2015visual} defined the lower bound EAO of state-of-the-art tracker as 0.251. SATA tracker shows state-of-the-art performance according to this definition.

\begin{table}
\small
\begin{center}
\begin{tabular}{l|c|c|c|c}
\hline
  & SATA & CCOT & MDNet & SiamFC \\
\hline
EAO & 0.319 & 0.331 & 0.257 & 0.277 \\
Accuracy & 0.544 & 0.539 & 0.541 & 0.549 \\
Robustness & 0.243 & 0.238 & 0.337 & 0.382\\
\hline
\end{tabular}
\end{center}
\caption{Comparison with the state-of-the-art trackers on the VOT 2016 dataset. The results are evaluated with expected average overlap (EAO), accuracy (mean overlap) and robustness (average number of failures).}
\label{tab.vot2016}
\vspace{-4mm}
\end{table}

\subsection{Results on LaSOT}
To further validate SATA on a larger and more challenging scenario in the wild, we conduct experiments on LaSOT without any parameters fine-tune \cite{fan2019lasot}. The LaSOT dataset provides a large-scale annotations with 1,400 sequences and more 3.52 millions frames in the testing set. Table \ref{tab.lasot} reports the results of SATA on LaSOT testing set with protocol I. Without bells and whistles, the SATA achieves competitive results compared to recent outstanding trackers. Specifically, SATA increases AUC of success rate and precision significantly comoared to the baseline (SATA No agg).

\begin{table}
\small
\begin{center}
\begin{tabular}{l|c|c|c|c}
\hline
  & SATA & ECO & ECO\_HC & SATA (No agg) \\
\hline
Success rate & 0.331 & 0.324 & 0.304 & 0.300 \\
Precision & 0.344 & 0.338 & 0.320 & 0.311 \\
\hline
\end{tabular}
\end{center}
\caption{Comparison with the state-of-the-art trackers on the LaSOT dataset. The results are evaluated with AUC score of success rate and precision as OTB.}
\label{tab.lasot}
\vspace{-2mm}
\end{table}

\subsection{Ablation study}
In this section, we conduct ablation study on OTB2013 to illustrate the effectiveness of proposed components.We first analyze the contributions of multi-spatial feature aggregation. Then we illustrate how the number of historical frames affects visual object tracking performance.

\begin{figure}[pt]
\begin{center}
\includegraphics[width=0.98\linewidth]{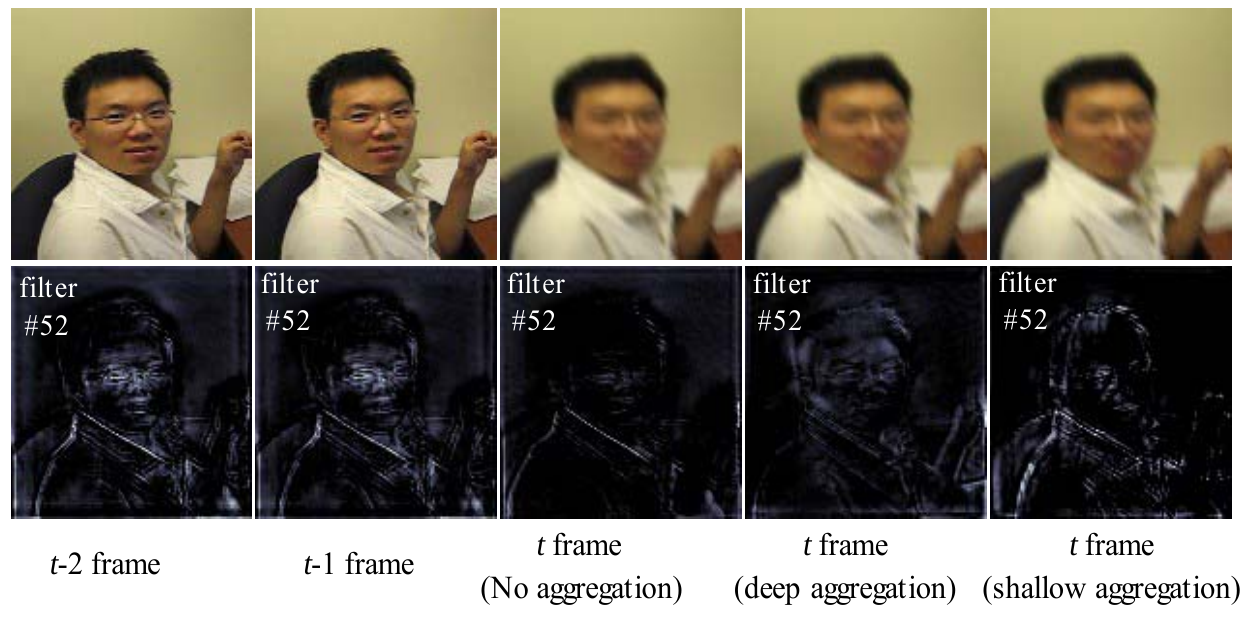}
\end{center}
\caption{Feature maps of “BlurFace” in OTB. Activations are high at frames \textit{t-2} and \textit{t-1} but low at frame \textit{t} due to motion blur. After aggregation, the feature map at the \textit{t} frame is improved.}
\label{fig:filtervisulization}
\end{figure}

\noindent
\textbf{Multi-scale feature aggregation analysis}. To verify the superiority of proposed multi-scale feature aggregation strategy and to assess the contributions of feature aggregation on different layer in our algorithm, we implement and evaluate six variations of our approach. Two metric are used to evaluate the performance of different architectures,  AUC means area under curve (AUC) of each success plot and FPS stands for mean speed. At first, the baseline is implemented that no aggregation is utilized (denoted by No Agg). Then the influence of number of layer for feature aggregation are tested. Historical frame number for this ablation experiments is $T=3$. We present our results in Table \ref{tab.ablation}, where $P^{l}$, $P^{l-1}$ and $P^{l-2}$ represent aggregation layer in feature pyramid. The algorithm $P^{l} P^{l-1}$ Agg, which assembles two highest layer feature aggregation, gains the performance with more than 0.05 compared to the version of baseline(No Agg). However, the performance boosting is not obvious but along with speed decreasing when we used more aggregation feature layer ($P^{l} P^{l-1} P^{l-2}$ Agg gains 0.006 than $P^{l} P^{l-1}$ Agg). $P^{l} P^{l-1}$ Agg makes a trade-off between speed and performance. $P^{l} P^{l-1}$ Agg operate on real time speed 26 FPS and AUC 0.677 when set $ T=3 $ (last line in Table \ref{tab.ablation}). As shown in \ref{fig:filtervisulization}, the shallow aggregation could favor  fine-grained spatial details and the deep aggregation maintains semantic patterns, which demonstrates our coarse-to-fine fusion paradigm in spatial is important  to preserve the location  information and improve the activation for motion blur.

In a word, our comparison results show that the aggregation of lower-resolution and strong semantic feature maps could increase tracking robustness and performance.

\begin{table}
\small
\begin{center}
\begin{tabular}{l|c|c|c|c|c|c}
\hline
Method & $P^{l}$ & $P^{l-1}$ & $P^{l-2}$ & T & AUC & FPS \\
\hline
No Agg & $\times$ & $\times$ & $\times$ & 3 & 0.646 & 62\\
$P^{l}$ Agg & \checkmark & $\times$ & $\times$ & 3 & 0.653 & 26 \\
$P^{l-1}$ Agg & $\times$ & \checkmark & $\times$ & 3 & 0.671 & 24 \\
$P^{l-2}$ Agg & $\times$ & $\times$ & \checkmark & 3 & 0.662 & 21 \\
$P^{l} P^{l-1}$ Agg  & \checkmark & \checkmark & $\times$ & 3 & 0.692 & 19  \\
$P^{l} P^{l-1} P^{l-2}$ Agg & \checkmark & \checkmark & \checkmark & 3 & 0.698 & 14 \\
$P^{l} P^{l-1} $ Agg & \checkmark & \checkmark & \checkmark & 2 & 0.677 & 26 \\
\hline
\end{tabular}
\end{center}
\caption{Configurations of ablation study.}
\label{tab.ablation}
\vspace{-2mm}
\end{table}

\begin{figure}[H]
\begin{center}
   \includegraphics[width=0.8\linewidth]{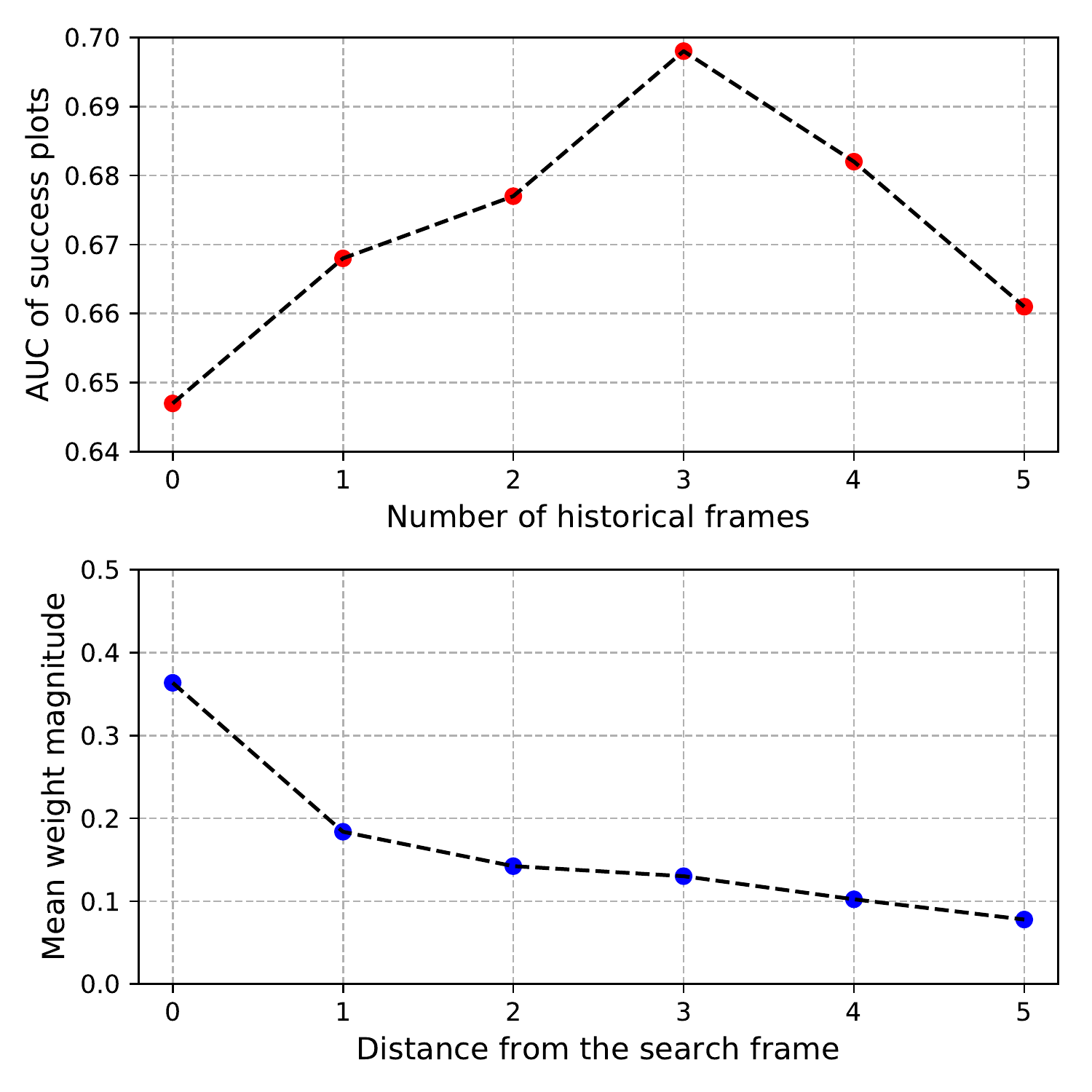}
\end{center}
\vspace{-2mm}
   \caption{A figure illustrating some of our ablation experiments. Top: we plot AUC of success plots as a function of the number of historical frames. Bottom: The contribution of each historical frames are represented with average weight magnitudes $w_{i-\tau \rightarrow i}$.}
\label{fig.weight_auc_frame}
\vspace{-2mm}
\end{figure}

\noindent
\textbf{Temporal feature aggregation analysis}. In online tracking, complementary features are derived from previous frames of the search image with fixed window size $T$, and $T$ could influence the performance of tracking. More historical frames mean that tracker could aggregate more temporal and motion information,  but more historical frames will also introduce greater appearance changes. Then we need choose a appropriate number of historical frames to make a trade off. Figure \ref{fig.weight_auc_frame} (Top) illustrate how the number of historical frames affects tracking accuracy (AUC of success plots). We notice that AUC increases first and then decreases as we add more historical frames. We further explore the reason of performance degradation with aggregation weight analysis.
Figure \ref{fig.weight_auc_frame} (Bottom) shows average weight magnitudes $w_{i-\tau \rightarrow i}$ for different values of $\tau$. This weight magnitudes could indicate how similar each historical frame is to the search frame. We note that historical frames that are further away from the search frame have small weights, which means greater difference between historical and search frame in feature embedding space. Hence, an aggregation of long-distance historical frames may introduce appearance bias to search frame.

\section{Conclusion}
In this work, we propose a end-to-end spatial-aware Temporal Aggregation Network for Visual Object Tracking(SATA) which makes use of the rich information from multi-scale and variable-length temporal windows of historical frames. We developed a Aligned Module using deformable ConvNets to estimate the motion changing of multi-spatial resolution feature map across frames, and features of historical frame are sampled and aggregated to search frame with pixel-level alignment and multi-scale aggregation. In experiments, our method achieves leading performance in OTB2013, OTB2015, VOT2015, VOT2016 and LaSOT and operates at real time speed of 26 FPS.

{\small
\bibliographystyle{ieee}
\bibliography{egpaper_SATA}
}

\end{document}